\documentclass{ifacconf}

\usepackage{graphicx}      
\usepackage{natbib}        

\usepackage{graphics}
\usepackage[flushleft]{threeparttable}
\graphicspath{ {Figures/} }

\usepackage[dvipsnames]{xcolor}
\usepackage{url} 

\begin{document}
\begin{frontmatter}

\title{Evaluation of Lidar-based 3D SLAM algorithms in SubT environment\thanksref{footnoteinfo}} 

\thanks[footnoteinfo]{This work has been partially funded by the European Unions Horizon 2020 Research and Innovation Programme under the Grant Agreement No. 869379 illuMINEation. Corresponding author's e-mail: antkov@ltu.se)}

\author[First]{Anton Koval} 
\author[First]{Christoforos Kanellakis} 
\author[First]{George Nikolakopoulos}

\address[First]{The Authors are with the Robotics and AI Group, Department of Computer, Electrical and Space Engineering, Lule\r{a} University of Technology, Lule\r{a}, Sweden.}

\begin{abstract}                
Autonomous navigation of robots in harsh and GPS denied subterranean (SubT) environments with lack of natural or poor illumination is a challenging task that fosters the development of algorithms for pose estimation and mapping. Inspired by the need for real-life deployment of autonomous robots in such environments, this article presents an experimental comparative study of 3D SLAM algorithms. The study focuses on state-of-the-art Lidar SLAM algorithms with open-source implementation that are i) lidar-only like BLAM, LOAM, A-LOAM, ISC-LOAM and hdl\_graph\_slam, or ii) lidar-inertial like LeGO-LOAM, Cartographer, LIO-mapping and LIO-SAM. The evaluation of the methods is performed based on a dataset collected from the Boston Dynamics Spot robot equipped with 3D lidar Velodyne Puck Lite and IMU Vectornav VN-100, during a mission in an underground tunnel. In the evaluation process poses and 3D tunnel reconstructions from SLAM algorithms are compared against each other to find methods with most solid performance in terms of pose accuracy and map quality.
\end{abstract}

\begin{keyword}
SLAM, SubT, Robotic Systems, Lidar, Autonomy Package
\end{keyword}

\end{frontmatter}

\section{INTRODUCTION}
Simultaneous Localisation and Mapping (SLAM) is the challenging task which addresses the problem of autonomous robot navigation in unknown environment~\cite{sualeh2019simultaneous:11,agha2021nebula:29} during which robot incrementally acquires a map of the environment using on-board perception and inertial sensors, while trying to localise itself within this map. Robust pose estimation is a crucial task for a mobile robot control. In the lab environment it can be solved by the means of the motion capture system, like Vicon. But it is not always possible to deploy it to the target area of robot navigation. Thus, one of the goals for robots is to being able to perform autonomous navigation independently from the infrastructure~\cite{lindqvist2020nonlinear:30}. Outdoors, in the open areas it is possible to use Global Positioning System (GPS) for pose estimates. However, in GPS denied environments like subterranean areas this solution will not work or will require installation of additional equipment, which is not always applicable~\cite{mansouri2020deploying:31}. Currently, there exist SLAM algorithms that are based on the camera and ranging sensors~\cite{sualeh2019simultaneous:11}, which in general require fast mobile computers to deliver real-time map building and pose estimation.

Overtime the technological development of mobile CPUs, sensors' miniaturisation and long endurance batteries allowed to run SLAM algorithms onboard, which fostered bringing robotic applications into challenging SubT environments. In such areas the crucial factor is human safety, which can be improved by the means of robots that perform autonomous inspection and increase situational awareness of human workers about environment by providing its reconstruction. This imposes high requirements to the mapping quality and localisation accuracy.

In SubT environments with poor illumination visual SLAM methods~\cite{sualeh2019simultaneous:11} tend to demonstrate poor performance, which is not acceptable. In contrary to them the lidar-based methods~\cite{ren2019robust:15} can deliver a solid performance for pose estimation and map presentation of the environment. Nevertheless, their performance may degrade over the time due to peculiarities of the SubT environments like long featureless, self-similar tunnel areas, dusty tunnels and sensor limitations. 

In this article we will focus on the evaluation of the major 3D SLAM methods that are open source and compatible with Robot Operating System (ROS)~\cite{quigley2009ros:14}, which became state-of-the-art (SoA) frameworks for the robotic community. To date there are several articles that conducted a comparison of lidar-based SLAM algorithms. For example, in~\cite{milijas2021comparison:17} Cartographer, LOAM and hdl\_graph\_slam are compared in open outdoor environment. Also~\cite{cong2020mapping:16} presents the evaluation of the developed SLAM method against A-LOAM and LeGO-LOAM methods in outdoor environment. In~\cite{ren2019robust:15} the evaluation of four SLAM methods in indoor environments is shown. Authors in study~\cite{milijas2020comparison:27} are comparing only two methods. Finally, the comparative analysis in study~\cite{zou2021comparative:18} is lacking one of the most recent and advanced SLAM methods LIO-SAM~\cite{liosam2020shan:7} and Fast-LIO~\cite{xu2021fast:26}, while the evaluation took place in the warehouse area with lots of features. 

Thus in this study we will experimentally evaluate nine open-source ROS compatible 3D SLAM lidar-based algorithms in the underground environment.

The main contributions of this work are: (1) The evaluation of nine SoA Lidar based 3D SLAM methods using a SubT dataset to demonstrate their performance in such environments, motivated by the emerging need for their deployment in underground tunnel environments. More specifically, the evaluation dataset was collected during the exploration mission of the Boston Dynamics Spot along an underground area with multiple tunnels. The onboard sensor suite consisted of the Velodyne Puck Lite lidar that is a SoA sensor for autonomous navigation and the Vectornav vn-100 IMU, hardware that is commonly used in multiple robotic systems and highly relevant when it comes to SubT research efforts. (2) The quantitative and qualitative comparison of the pose estimation and produced 3D maps of the environment for all methods which will make easy for the robotic community to assess and understand their advantages and disadvantages, including the selection of a SLAM algorithmic framework for this applications. 

The rest of the article is organised as follows. Section 2 introduces the SLAM methods, Section 3 follows their evaluation and finally the article is summarised with Conclusion section.
\section{SLAM ALGORITHMS}
In this study it have been selected all the major SLAM algorithms that have real-time operation and use as an input a point cloud from a 3D lidar or couple it with IMU measurements. This selection introduces firstly the robotic platform, secondly the SubT environment and finally the SLAM methods. Worth noting that not all of them have an article that could be cited and that introduces and explains the method, however they remain to be operational and can stand as a baseline for numerous applications and future developments towards performance improvements and multi-agent SLAM~\cite{mourikis2006predicting:19,krinkin2017modern:20}. All the selected methods were tested with ROS Melodic and Ubuntu 18.04. 
\subsection{Robotic platform and data set} \label{dataset}
\begin{figure}[htbp!]
  \centering
  \includegraphics[width=0.65\linewidth]{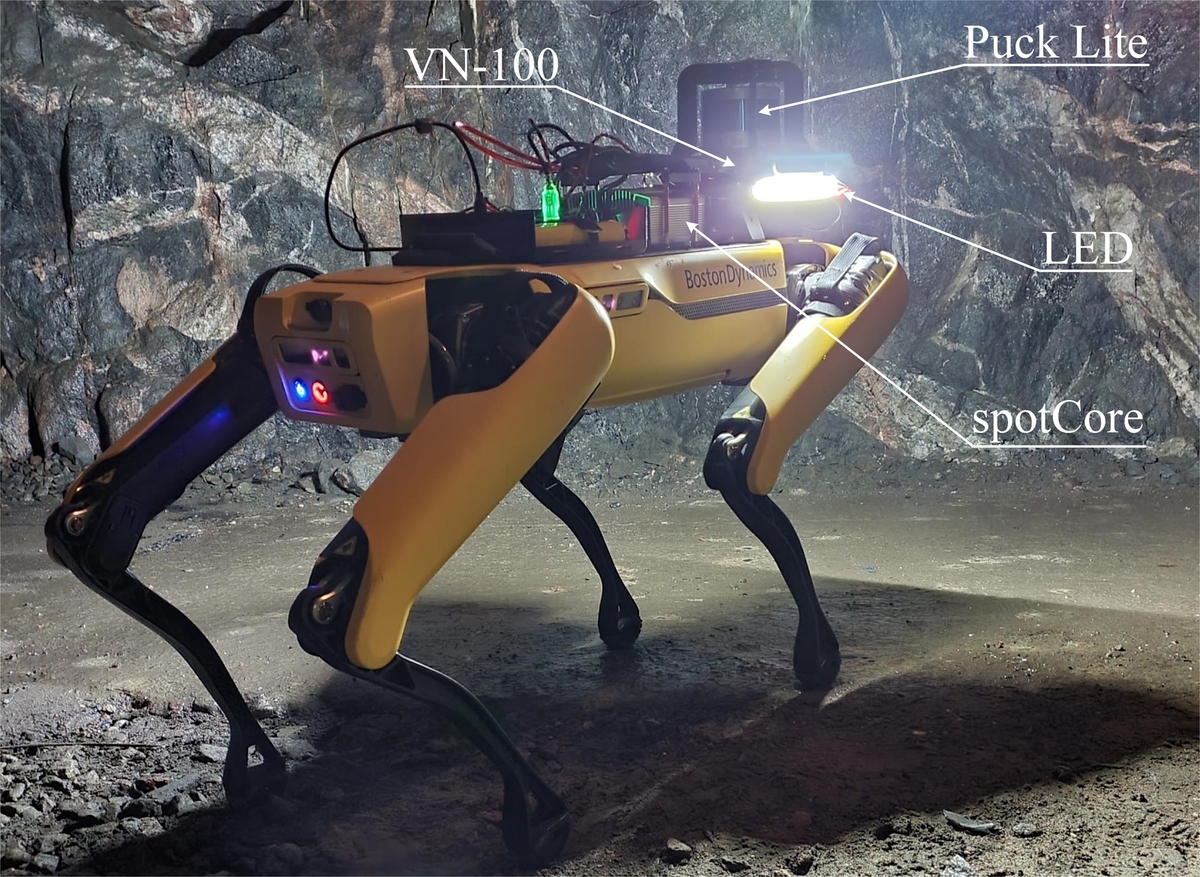}
  \vspace*{-2mm}
  \caption{Spot robot equipped with sensors for dataset collection in the underground tunnel.}
  \label{fig:spot}        
\end{figure}

As robotic platform for data collection was used Spot legged platform~\cite{koval2022experimental:28} developed by Boston Dynamics (Figure~\ref{fig:spot}). This robot is capable of moving with the velocity of up to 1.6 m/s, carrying up to 14 kg of payload and traversing challenging terrains. On top of Spot was placed the payload that includes 3D lidar, IMU, spotCore and batteries as depicted on the Figure~\ref{fig:spot}. The onboard computer has Intel Core i5 CPU with 16GB of RAM with Ubuntu 18.04 and ROS Melodic. To provide an unobstructed field of view for 3D lidar it was mounted as a column structure in a front looking position with separate LED light bars Lustreon DV12V 10W with dimensions $170 \times 15$ mm pointing towards front, left and right.

\begin{figure}[htbp!]
  \centering
  \includegraphics[width=0.81\linewidth]{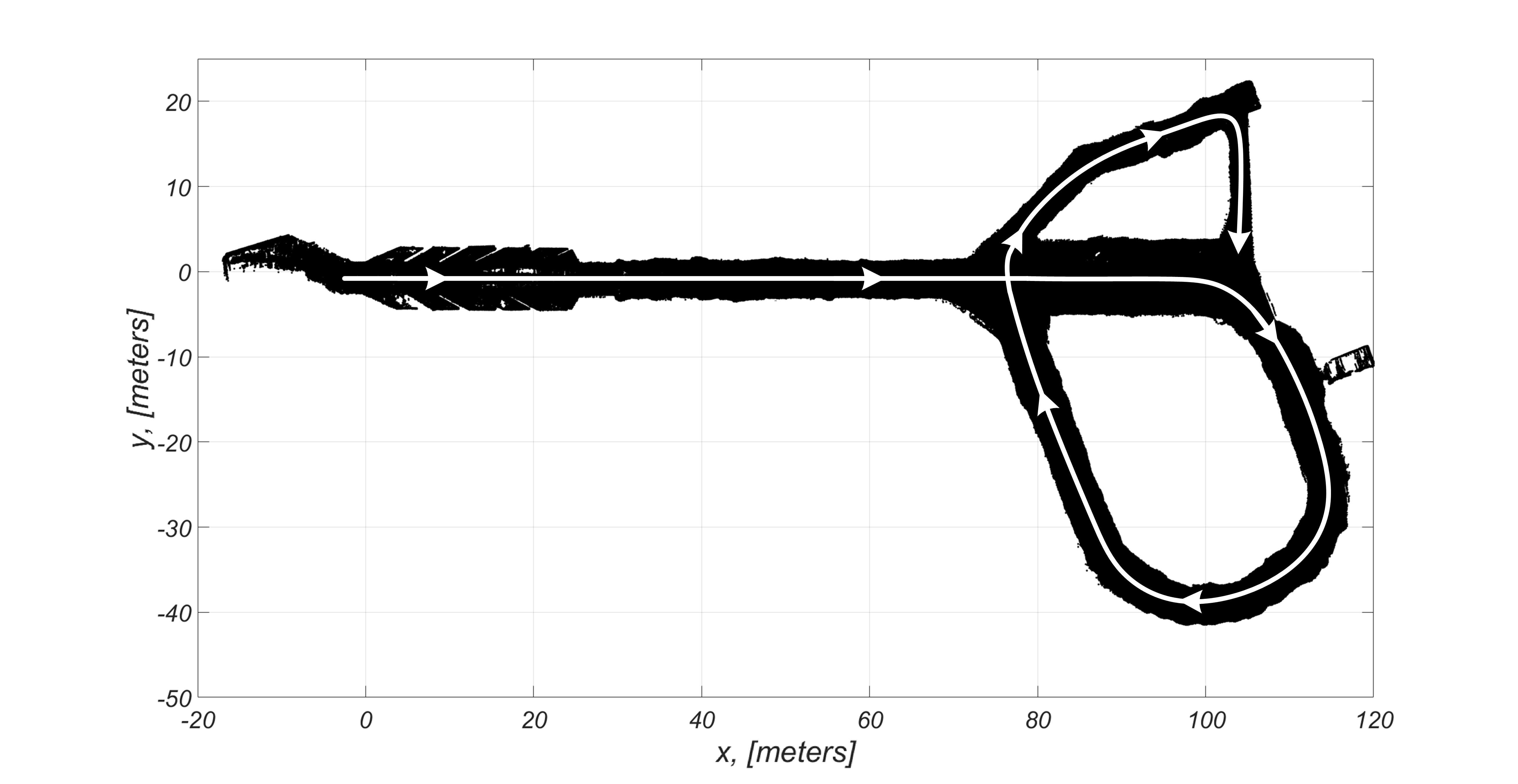} 
  \vspace*{-2mm}
  \caption{A top view of the map generated from the 3D lidar scans from the field test environment. A solid line represents the traverse through the tunnel, while arrows depict the direction of traverse.}
  \label{fig:tunnel-map}        
\end{figure}

The dataset was collected from Lule\aa~ Sweden underground tunnel with manually controlled Spot as shown on the Figure~\ref{fig:tunnel-map}. It was recorded in one pass storing the measurements from IMU and 3D lidar in a ROS bag file~\footnote{\url{http://wiki.ros.org/rosbag}}. The top view of a 3D reconstructed map is depicted on the Figure~\ref{fig:tunnel-map}. 
The data were collected with the sensor configuration as depicted on the Figure~\ref{fig:spot} with IMU publishing rate set to 200Hz and 3D lidar publishing rate set to 10Hz. 
Obtaining a ground truth in SubT environment is a challenging task, thus in this study only relative comparison of the algorithms is provided.

In the following subsections we will briefly introduce the selected algorithms. 

\subsection{B(erkeley) L(ocalization) A(nd) M(apping)}
Berkeley localization and mapping (BLAM)~\footnote{\url{https://github.com/erik-nelson/blam}} is an open-source ROS package for lidar graph‐based real-time localization and mapping. It computes loop closures by iterative closest point (ICP) scan matching using scans from nearby poses. For map optimisation it uses Georgia tech smoothing and mapping (GTSAM) library. BLAM is capable of building very dense and precise maps online which makes it a computationally expensive method~\cite{nava2019visual:21}.
\subsection{Laser Odometry and Mapping methods}
Laser Odometry and Mapping (LOAM)~\cite{zhang2014loam:2,loam_velodyne_2016:1} or loam\_velodyne is a real-time method that is able simultaneously estimate odometry and build a map using 3D lidar. This method solves the SLAM task by splitting it into two algorithms. One for computing odometry and second for incremental map building, additionally it also estimates velocity of the lidar. LOAM doesn't have a loop closure, which prevents it to recognise previously visited areas, instead it implements feature point matching which allows to ensure fast odometry computation and accurate map building. The method has IMU support which allows to obtain higher accuracy in comparison with using only lidar.

A-LOAM~\cite{zhang2014loam:2} is an advanced implementation of LOAM~\cite{zhang2014loam:2}, which uses the Eigen library for linear algebra operations and Ceres Solver for solving the corresponding optimisation problem.

Fast LiDAR Odometry and Mapping (F-LOAM)~\cite{wang2021f:25} is an optimized version of LOAM and A-LOAM which is based on a  a non-iterative two-stage distortion compensation method that allows to lower the computation time. F-LOAM combines feature extraction, distortion compensation, pose optimization, and mapping. 

\subsection{ISC-LOAM}
Intensity Scan Context based Full SLAM Implementation (ISC-LOAM)~\cite{wang2020intensity:4} is another algorithm designed for 3D lidars. It combines a global descriptor that incorporates geometry and intensity characteristics. The proposed loop closure detection approach is based on a two-stage hierarchical intensity scan context (ISC) for place recognition, which allows to improve computational performance. The ISC incorporates fast binary-operation based geometry indexing and intensity structure re-identification.
\subsection{hdl\_graph\_slam}
hdl\_graph\_slam~\cite{koide2018portable:3} is an open source ROS package for real-time simultaneous localization and mapping with a 3D lidar. This method is based on the pose graph SLAM in which loop closure detection is based on the Normal Distributions Transform (NDT) scan matching between the consecutive frames. The NDT method has better scan matching performance in the applications with 3D lidars than other algorithms. In it the Unscented Kalman Filter is used for pose estimation.

Complementary to the developed package is the \linebreak  \textit{hdl\_localization} module that implements relocalization on the known map. In literature this problem can be referred as global localization problem or the kidnapped robot problem~\cite{se2002global:24}.
%
\subsection{LeGO-LOAM}
Lightweight and ground optimized lidar odometry and mapping (LeGO-LOAM)~\cite{legoloam2018:5} is a real-time method designed for pose estimation mapping with unmanned ground vehicles in complex environments with changing terrain. It leverages ground separation by performing point cloud segmentation, which allows to reject points that can represent unreliable features. LeGO-LOAM performs pose estimation using two-step optimisation. During the first step, planar features are extracted from the ground to obtain $z$, roll, pitch and during the second step the remained $x$, $y$, yaw are obtained by matching features extracted from the point cloud. This method also supports loop closure, which is implemented using ICP. 
\subsection{Cartographer}
Cartographer~\cite{hess2016real:22}~\footnote{\url{https://github.com/cartographer-project/cartographer\_ros}} is a system developed by Google for real-time simultaneous localization and mapping for 2D and 3D. It has support of distinctive sensor configurations. For 3D SLAM it requires to have IMU data that are needed for the initial guess for determining the orientation of the lidar scans. At the best pose estimate it take the scans from a lidar and translates them into a probability grid which is used to build a submap. The recently finished submaps and scans are considered for loop closure through the scan matching. The scan matching relies on a branch-and-bound algorithm. Cartographer combines separately local and global SLAM approaches. In the local SLAM the Ceres matcher is used to find poses that optimally match the submap. This process slowly accumulates error, which is eliminated by the loop-closure mechanism that is based on the Sparse Pose Adjustment (SPA)~\cite{konolige2010efficient:23}.
\subsection{LIO-mapping}
A Tightly Coupled 3D Lidar and Inertial Odometry and Mapping (LIO-mapping)~\cite{ye2019tightly:6} is a real-time method for 3D pose estimation and mapping. In this method IMU is tightly coupled with lidar in order to jointly minimize the cost derived from lidar and IMU measurements. This method uses the sliding window approach to limit the number of computation by including new pose estimates and deprecating oldest ones in the window. However, LIO-mapping remains to be computationally expensive method for real-time navigation taking more than 0.2 seconds for simultaneous odometry estimation and mapping with a 16 line 3D lidar.
\subsection{LIO-SAM}
LIO-SAM~\cite{liosam2020shan:7} is a real-time tightly-coupled lidar-inertial odometry package, which built from LeGo-LOAM and is an ICP-based method. This method is constructed as a factor graph, which makes it easy to incorporate additional sensors. In its implementation LIO-SAM adds IMU preintegration in an incremental smoothing and mapping approach with the Bayes tree. Worth noting that this method is capable of processing data 13 times faster than real-time.

\subsection{FAST-LIO}
FAST-LIO~\cite{xu2021fast:26} is a LiDAR-inertial odometry framework in which a tightly-coupled iterated Kalman filter is used to fuse LiDAR feature points with
IMU measurements. In this method planar and edge features are extracted from the lidar point cloud, at the next step these features together with IMU measurements are used for state estimation, after that the estimated pose is used to register the feature points into the global frame and to update the global map.

The global frame of FAST-LIO is defined as the first IMU's frame. That means the IMU's $x$-$y$-$z$ axis at the very beginning will be the global frame's $x$-$y$-$z$ axis. So the robot's orientation may not looks like you imagine.

\begin{table*}[t]
\renewcommand{\arraystretch}{0.92}
\vspace{-5pt}
\centering
\caption{SLAM packages$^1$}
\label{tab:slam-methods}
\begin{threeparttable}
\begin{tabular}{l|c|c|c|c|c|c} 
\hline
\multicolumn{1}{|l|}{SLAM package} & \multicolumn{1}{l|}{3D lidar} & \multicolumn{1}{l|}{IMU} & \multicolumn{1}{l|}{Loop closure} & \multicolumn{1}{l|}{Real-time 3D} & \multicolumn{1}{l|}{Real-time} & \multicolumn{1}{l|}{Relocalization} \\
\multicolumn{1}{|l|}{}             & \multicolumn{1}{l|}{}         & \multicolumn{1}{l|}{}    & \multicolumn{1}{l|}{}             & \multicolumn{1}{l|}{point cloud map}  & \multicolumn{1}{l|}{operation} & \multicolumn{1}{l|}{on a known map} \\ 
\hline
BLAM                               & R                             & N                        & Yes (ICP)                           & Yes                                 & Yes                  & No                  \\
[0.25em]
loam\_velodyne                     & R                             & O                        & No                                 & Yes                                 & Yes                   & No            \\
[0.25em]
A-LOAM                             & R                             & N                        & No                                 & Yes                                 & Yes                   & No           \\
[0.25em]
F-LOAM                             & R                             & N                        & No                                 & Yes                                 & Yes                   & No           \\
[0.25em]
ISC-LOAM                           & R                             & N                        & Yes (ISC)                           & Yes                                 & Yes                  & No            \\
[0.25em]
hdl\_graph\_slam                   & R                             & O                        & Yes (NDT)                                 & Yes                                  & Yes                      & Yes$^2$         \\
[0.25em]
LeGO-LOAM                          & R                             & O                        & Yes (ICP)                           & Yes                                 & Yes                  & No            \\
[0.25em]
Cartographer                       & R                             & R                        & Yes (SPA)                           & No                                 & Yes                   & Yes           \\
[0.25em]
LIO-mapping                        & R                             & R                        & No                                  & Yes                                 & Yes                      & No        \\
[0.25em]
LIO-SAM                            & R                             & R                        & Yes (ICP)                                  & Yes                                 & Yes                      & No         \\
[0.25em]
Fast-LIO                            & R                             & R                        & No                                  & Yes                                 & Yes                      & Yes$^3$         \\
[0.25em]
\hline
\end{tabular}
\begin{tablenotes}
        \small
        \item $^1$~In 3D lidar and IMU columns, R - required, N - not required, O - operational.
        \item $^2$~Relocalization on the known map is implemented in \textit{hdl\_localization} package.
        \item $^3$~Relocalization on the known map is implemented in \textit{FAST\_LIO\_LOCALIZATION} package.
\end{tablenotes}
\end{threeparttable}
\end{table*}

All the SLAM methods, required hardware and their features are summarised in the Table~\ref{tab:slam-methods}

The next step was to run the recorded data on the SLAM methods.
%
\section{EXPERIMENTAL EVALUATION AND DISCUSSIONS}
In this Section the evaluation of the SLAM methods is performed in the SubT environment based on the collected dataset introduced in the subsection~\ref{dataset}.  

The benchmarking of SLAM methods was conducted on the computer with Intel i7 9th generation CPU, 64 GB of RAM, Ubuntu 18.04 and ROS Melodic. For the algorithms' comparison we have done our best efforts to tune all methods with respect to the given hardware configuration of the autonomy package.

\subsection{Evaluation and comparison of trajectories}
\begin{figure}[htbp!]
  \centering
  \includegraphics[width=0.85\linewidth]{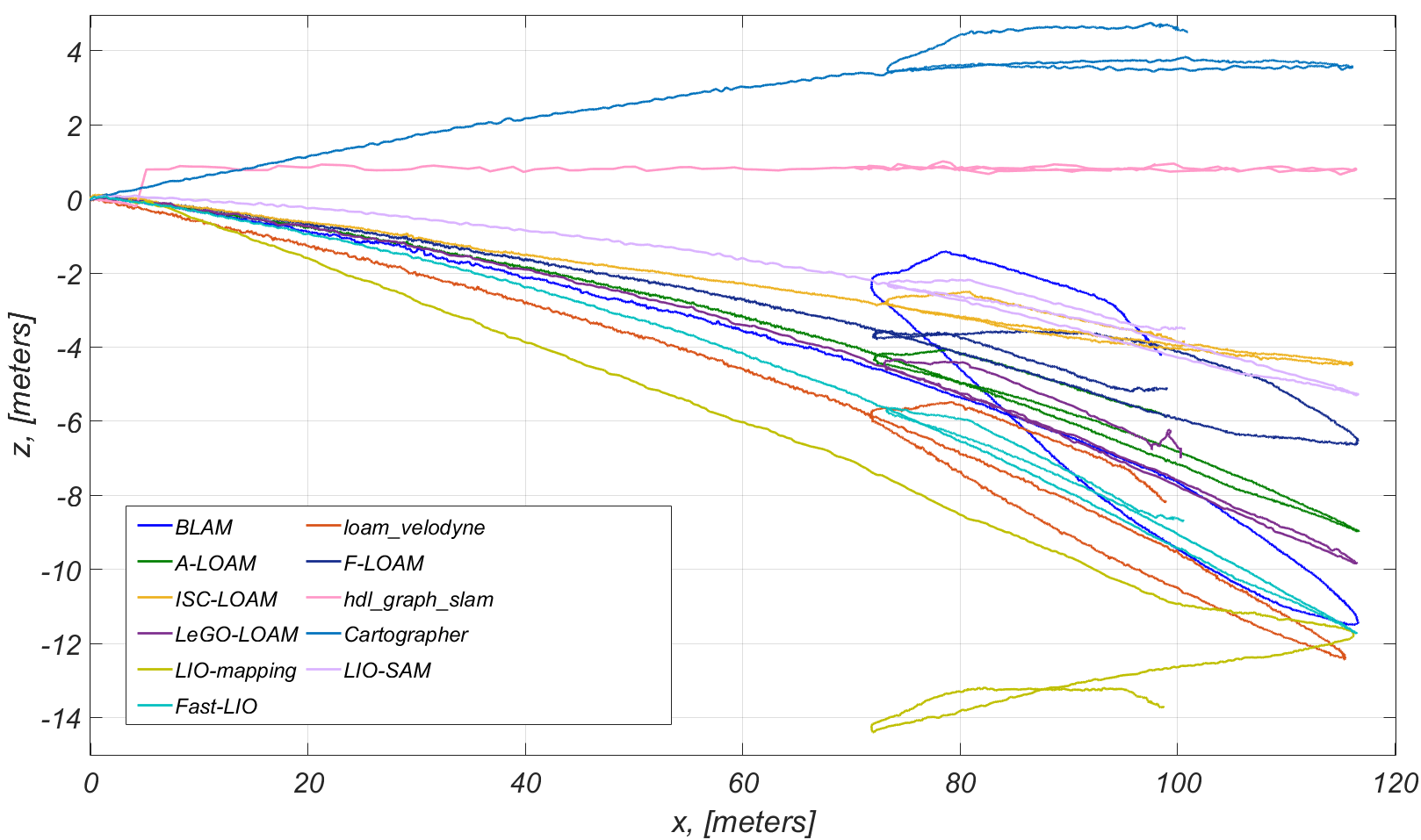} 
  \vspace*{-2mm}
  \caption{Estimation accuracy of $z$ coordinate for all methods}
  \label{fig:z-axis}        
\end{figure}

The collected dataset from the underground tunnel represents a challenging environment for lidar-based SLAM algorithms in terms of lack of features, repetitiveness and narrow size, while additionally two close-loop branches will allow to evaluate methods' loop closure performance. 
All SLAM methods were thoroughly tuned with our best efforts, however, based on multiple evaluations we came to conclusion that the field of view of the VLP16 Lite is not sufficient to capture enough data in vertical dimension, which leads to high uncertainty about $z$ axis, which cannot be compensated even with the use of IMU, as depicted in the Figure~\ref{fig:z-axis}.

Based on our slope measurements with DeWALT laser level going for 50 meters from the entrance towards the center of the tunnel we obtained that the tunnel has a positive slope with angle of incline equal to 0.95 degrees. However, none of the considered methods, was able to estimate the angle of inclination correctly even at the straight part of the tunnel, as such for Cartographer the inclination angle is estimated about 2.91 degrees, for hdl\_graph\_slam it is about 0.1 degrees, for LIO-SAM it is about -1.4 degrees and for Fast-LIO about -3.7 degrees. Worth noting that all methods except Cartographer and hdl\_graph\_slam estimated negative slope, which can mean that the remained methods that fuse IMU data were strongly relying on lidar sensor rather then on IMU.

\begin{figure}[htbp!]
  \centering
  \includegraphics[width=0.85\linewidth]{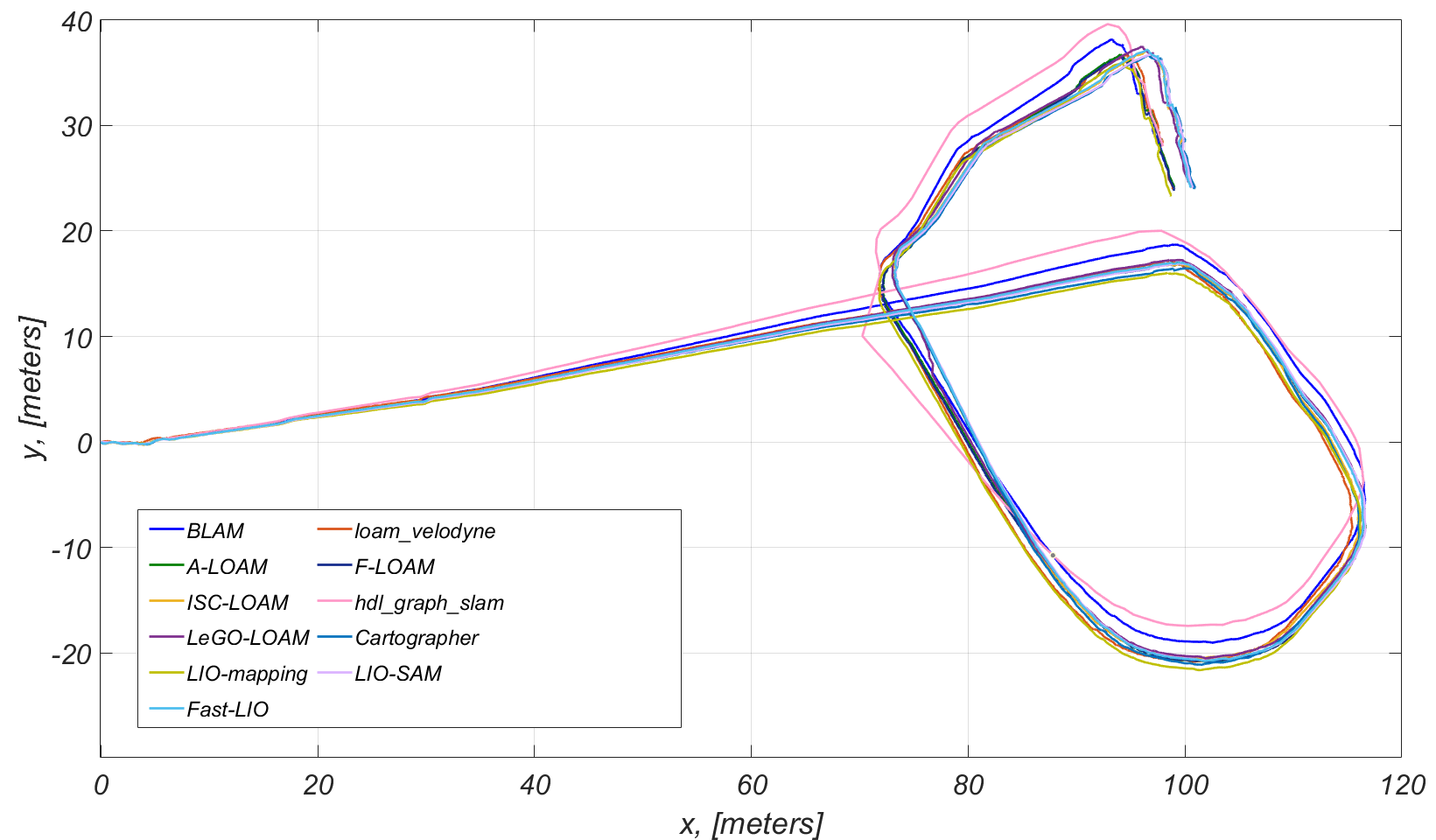} 
  \vspace*{-2mm}
  \caption{2D pose estimation accuracy for all methods}
  \label{fig:xy-axis}        
\end{figure}

Alike to the vertical dimension, in the horizontal dimension VLP16 Lite has high data redundancy, which allows to all the methods to cope with the $x$ and $y$ 2D pose estimation, as depicted in Figure~\ref{fig:xy-axis}.

\begin{figure*}[ht!]
  \centering
  \includegraphics[width=0.88\linewidth]{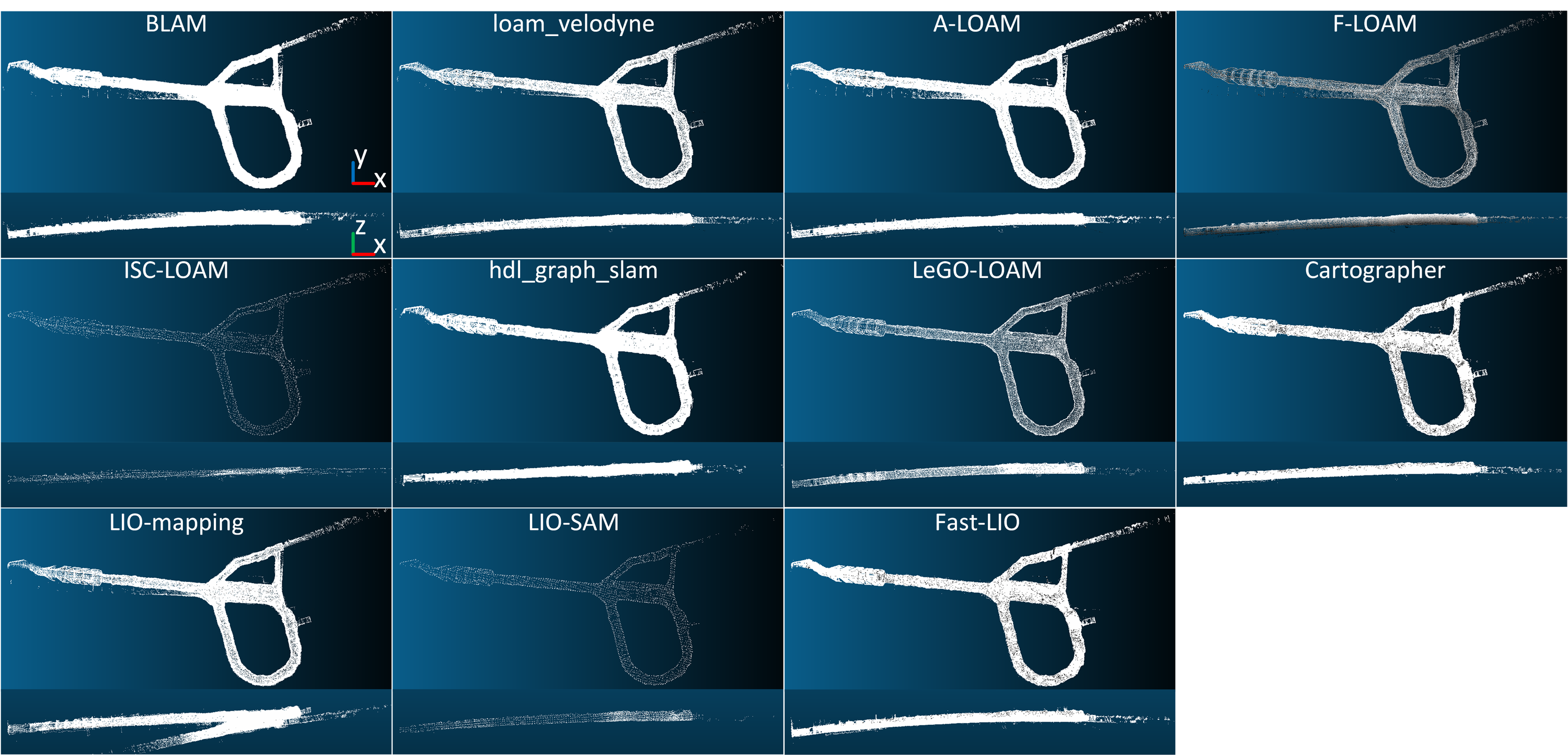} 
  \vspace*{-2mm}
  \caption{Produced 3D point cloud maps from each method. The global coordinate frame for all figures is as defined for BLAM method, in which on the upper part is shown $xy$ view and on the bottom $xz$ view.}
  \label{fig:slam-maps}        
\end{figure*}

As it can be seen from the Figure~\ref{fig:tunnel-map}, the 2D pose estimated by all SLAM methods is inline with the map of the tunnel. In overall, all algorithms were able to estimate thepo e along the tunnel with hdl\_graph\_slam as a clear outlier. We will continue this analysis in the next subsection, where we will evaluate the produced maps. For relative methods' evaluation we've calculated and compared the total travelled distance, which is shown in the Table~\ref{tab:slam-distance} and calculated confidence interval of $[251.20\; 256.66]$ meters with confidence level of 95\%. From this can be seen that BLAM, A-LOAM, LeGO-LOAM, Cartographer and Fast-LIO distance measurements have more trusted values.

\begin{table}[h]
\renewcommand{\arraystretch}{0.92}
\centering
\caption{Summary of the comparison}
\label{tab:slam-distance}
\begin{threeparttable}
\begin{tabular}{l|c|c} 
\hline
\multicolumn{1}{|l|}{SLAM package} & \multicolumn{1}{|c|}{Travelled distance,} & \multicolumn{1}{c|}{Number of points} \\
\multicolumn{1}{|l|}{}             & \multicolumn{1}{|c|}{[meters]}             &\multicolumn{1}{c|}{in the produced map}         \\ 
\hline
BLAM                               & 253.6177    &   2 429 696                      \\
[0.25em]
loam\_velodyne                     & 257.6209    &   161 766                     \\
[0.25em]
A-LOAM                             & 255.1464    &   274 430                       \\
[0.25em]
F-LOAM                             & 258.6841    &   60 429                     \\
[0.25em]
ISC-LOAM                           & 260.2869    &   4 586                      \\
[0.25em]
hdl\_graph\_slam                   & 249.2185    &   1 600 192                     \\
[0.25em]
LeGO-LOAM                          & 252.9649    &   37 481                     \\
[0.25em]
Cartographer                       & 254.4119    &   8 232 813                    \\
[0.25em]
LIO-mapping                        & 247.5393    &   166 977                    \\
[0.25em]
LIO-SAM                            & 249.2190    &   5 594 126                      \\
[0.25em]
Fast-LIO                           & 254.5439    &   15 675 216                     \\
[0.25em]
\hline
\end{tabular}
\end{threeparttable}
\end{table}

\subsection{Evaluation and comparison of point clouds}
The mapping results allow to see the clear difference between methods with and without loop closure, which might be not that noticeable from the positioning results. In Figure~\ref{fig:slam-maps} the produced maps from each algorithm are depicted. Based on it, one can say that the lack of loop closure in loam\_velodyne, A-LOAM, F-LOAM and LIO-mapping is resulting in a point cloud map that is misaligned. In line with this group of methods Fast-LIO, though does not have a loop closure feature, could efficiently deal with the environment and produce a correct map. 

Among methods with implemented a loop closure capability, one can say that all of them except BLAM can provide a correct map. The hdl\_graph\_slam, while it is capable to produce the correct map, it remains to be outlier in pose estimation, though this finding could be due to a potential imperfection of our tuning. Depending on the point cloud registration approach, in default configuration, SLAM methods produce the following number of points, as shown in the Table~\ref{tab:slam-distance}, from which we can see that the least memory intensive method for storing the map is ISC-LOAM and Fast-LIO is the most memory expensive. Though, the overall number of points in the map can be reduced by additional method configuration, as for example in the Figure~\ref{fig:slam-maps}, the map size for LIO-SAM method is reduced to 4820 points.

\begin{figure}[htbp!]
  \centering
  \includegraphics[width=0.85\linewidth]{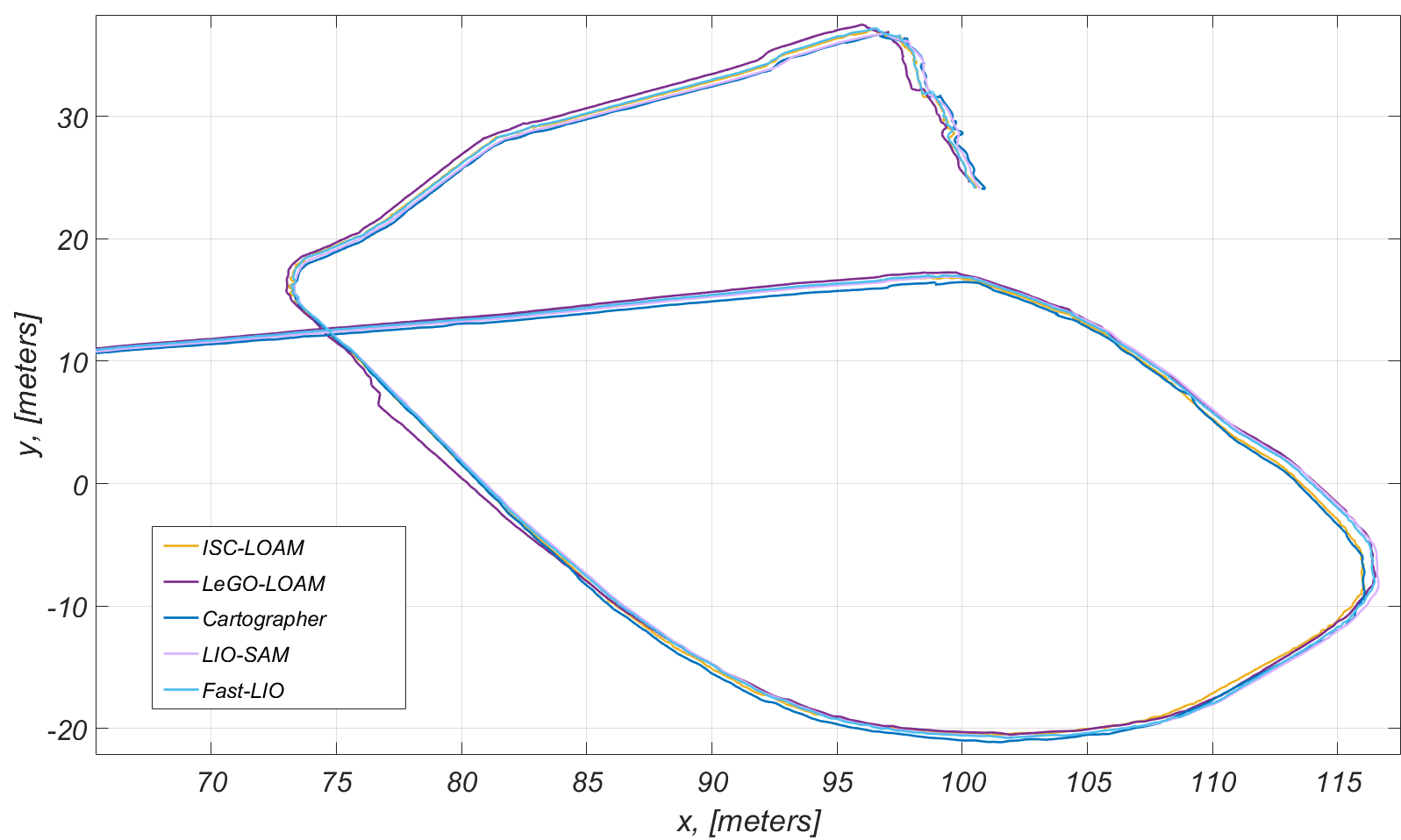} 
  \vspace*{-2mm}
  \caption{2D pose estimation accuracy for the selected methods}
  \label{fig:xy-axis-loop-closure}        
\end{figure}

Thus, if we replot the Figure~\ref{fig:xy-axis} at the junction area of the tunnel and leave on it only methods that provided solid results in pose estimation and mapping, as shown in the Figure~\ref{fig:xy-axis-loop-closure}, one can say that all methods demonstrate good stability and performance. However, as it can be noted in the junction area LeGO-LOAM performed loop closure so the map was corrected, but not the trajectory. Worth noting that IMU impact on the method performance, seeing that Fast-LIO is performing equally with methods with loop closure.

%

\section{Conclusion}

In this article we compared the most recent and SoA lidar SLAM methods' performance in a demanding SubT environment. For that we have collected the dataset using a Spot robot equipped with our autonomy package and configured all SLAM methods respectively. Based on the SLAM methods comparisons carried out in this article, it can be concluded that the equipped 3D lidar is not sufficient for 3D pose estimation leading to significant drift in $z$ axis due to the lack of features. Thus, our analysis was focused on 2D pose comparison. The evaluation results demonstrated that BLAM, A-LOAM, LeGO-LOAM, Cartographer and Fast-LIO produce more trusted results than other methods. ISC-LOAM is the least memory expensive method for map storing, though the produced map is very sparse comparing it with Fast-LIO. Moreover, it can be concluded that fusing IMU with lidar is helpful for correcting the pose estimation.

As a future direction we aim to use lidar sensor with wider field of view and to perform evaluation in a larger environment and with an uneven terrain that is challenging for legged robots.

\bibliography{ifacconf}
\end{document}